\documentclass{article}

\usepackage[preprint]{neurips_2025}

\usepackage[utf8]{inputenc} 
\usepackage[T1]{fontenc}    
\usepackage{hyperref}       
\usepackage{url}            
\usepackage{booktabs}       
\usepackage{amsfonts}       
\usepackage{nicefrac}       
\usepackage{microtype}      
\usepackage{xcolor}         
\usepackage{multirow}
\usepackage{amsmath}
\usepackage{algorithm}
\usepackage{xcolor}         
\usepackage{booktabs} 
\usepackage{multirow} 
\usepackage{makecell}
\usepackage{pifont} 
\usepackage{circledsteps}
\usepackage{makecell}
\usepackage{tabularx}
\usepackage[accsupp]{axessibility}  
\usepackage{graphicx}
\usepackage{amssymb}
\usepackage{enumitem}
\usepackage{subcaption}
\usepackage{bbding}
\usepackage[accsupp]{axessibility} %
\usepackage{algpseudocode}
\usepackage{algorithm}
\usepackage{nicefrac}
\usepackage{booktabs}
\usepackage{multirow}
\usepackage{multicol}
\usepackage{makecell}
\usepackage{colortbl}
\usepackage{arydshln}
\usepackage{subcaption}
\usepackage{amssymb}
\usepackage{pifont}
\usepackage{setspace}
\usepackage{tcolorbox}
\usepackage{natbib}
\usepackage{wrapfig}
\usepackage{gradient-text}
\usepackage{xcolor}
\usepackage{tikz}

\setcitestyle{numbers,square}
\tcbuselibrary{listings,breakable}

\definecolor{myred}{RGB}{200,50,50}

\newcommand{\eg}{\emph{e.g.}{}}
\newcommand{\ie}{\emph{i.e.}{}}
\newcommand{\etc}{\emph{etc}{}}
\newcommand{\cf}{\emph{cf. }{}}

\newcommand{\MethodName}{SyncLoop}

\definecolor{crosscolor}{rgb}{0.969,0.580,0.114} %
\definecolor{checkcolor}{rgb}{0.485,0.640,0.204} %

\def\cmark{{\color{checkcolor}\ding{52}}}
\def\xmark{{\color{crosscolor}\ding{56}}}

\usepackage{array} 
\newcolumntype{x}[1]{>{\centering\arraybackslash}p{#1pt}}

\newlength\savewidth

\usepackage{xcolor}
\definecolor{BrightRed}{RGB}{220,40,30}      
\definecolor{BrightOrange}{RGB}{255,140,0}   
\definecolor{Gold}{RGB}{255,200,0}           
\definecolor{Emerald}{RGB}{0,160,80}         
\definecolor{Cobalt}{RGB}{0,120,215}         
\definecolor{Sapphire}{RGB}{90,60,170}       
\definecolor{Amethyst}{RGB}{160,80,200}      

\definecolor{mycustomcolor}{HTML}{DDEEFF} 

\title{SyncLoop: A Multimodal Dual-Loop Framework for Self-Improving Mathematical Reasoning}

\author{
{\bf Xiuwei Chen$^1$}, ~
{\bf Wentao Hu$^2$,} ~
{\bf Hanhui Li$^1$,} ~
{\bf Yongxin Wang$^3$,} ~
{\bf Jun Zhou$^1$, } ~
\\
{\bf Zisheng Chen$^1$,} ~
{\bf Meng Cao$^3$,} ~
{\bf Yihan Zeng$^4$,} ~
{\bf Kui Zhang$^4$,} ~
{\bf Yu-jie Yuan$^4$,} ~
\\
{\bf Jianhua Han$^5$,}
{\bf Hang Xu$^5$,} ~
{\bf Xiaodan Liang$^{1}$}\thanks{Corresponding author.}
\\
\\
$^{1}$Sun Yat-sen University 
$^{2}$The Hong Kong Polytechnic University \\
$^{3}$MBZUAI
$^{4}$Huawei Noah's Ark Lab 
$^{5}$Yinwang Intelligent Technology Co. Ltd.
}

\begin{document}

\maketitle

\renewcommand\thefigure{\arabic{figure}}     
\renewcommand\thetable{\arabic{table}}        
\renewcommand\thesection{\arabic{section}}    
\renewcommand\theequation{\arabic{equation}}  

\begin{abstract}

{Recent advances in multimodal large language models (MLLMs) have shown impressive reasoning capabilities. However, further enhancing existing MLLMs necessitates high-quality vision-language datasets with carefully curated task complexities, which are both costly and challenging to scale. Although recent self-improving models that iteratively refine themselves offer a feasible solution, they still suffer from two core challenges: (i) most existing methods augment visual or textual data separately, resulting in discrepancies in data complexity (e.g., over-simplified diagrams paired with redundant textual descriptions); and (ii) the evolution of data and models is also separated, leading to scenarios where models are exposed to tasks with mismatched difficulty levels. To address these issues, we propose {\MethodName}, an automatic, closed-loop self-improving framework that jointly evolves both training data and model capabilities. Specifically, given a base dataset and a base model, {\MethodName}  enhances them by a cross-modal data evolution loop and a data-model evolution loop. The former loop expands the base dataset by generating complex multimodal problems that combine structured textual sub-problems with iteratively specified geometric diagrams, while the latter loop adaptively selects the generated problems based on the performance of the base model, to conduct supervised fine-tuning and reinforcement learning alternately. Consequently, our method continuously refines its model and training data, and consistently obtains considerable performance gains across multiple mathematical reasoning benchmarks. Our code will be released\footnote{\url{https://github.com/chen-xw/C2-Evo}}.}

\end{abstract}

\section{Introduction}
\label{sec:intro}
Recent advancements in large language models (LLMs) have achieved remarkable progress in solving problems, including mathematics  \cite{cobbe2021training, hendrycks2021measuring}, coding  \cite{chen2021evaluating, gu2024cruxeval}, etc.
These capabilities are enabled by advanced strategies, including chain-of-thought prompting  \cite{wei2022chain}, tool-augmented reasoning  \cite{feng2025retool, Hu2024VisualSS, li2025imagine}, \etc.
In particular, OpenAI o1  \cite{openai2024o1systemcard} and Deepseek-R1  \cite{guo2025deepseek} have shown that reinforcement learning plays a critical role in aligning model outputs with desired behaviors by using structured reward signals derived from correctness, consistency, or human preference.
This mechanism has proven especially effective in eliciting nuanced self-verification and self-correction behavior in LLMs, thereby reinforcing the reliability and depth of their reasoning chains, particularly in mathematical and logical domains.

Despite these advances, achieving such strong reasoning performance remains heavily reliant on large-scale, high-quality, and {complexity-aligned} datasets. As task complexity increases, collecting suitable training data becomes significantly more costly and difficult, presenting a major bottleneck to further progress. This challenge has sparked growing interest in \textit{self-improving} paradigms, where models iteratively enhance their capabilities by 
{
generating new synthetic data and refining reasoning traces.}

Recent studies have shown that reasoning abilities can be substantially improved through carefully curated and progressively challenging data. 
For example, OpenVLThinker \cite{deng2025openvlthinker} adapts the self-improvement paradigm to the vision-language domain by iteratively alternating between supervised fine-tuning (SFT) and reinforcement learning (RL), distilling R1-style reasoning traces from text-based models into multimodal contexts.


\begin{wraptable}[7]{r}{0.48\linewidth}
    \centering
    \vspace{-1.1em}
    \setlength{\tabcolsep}{4pt}
    \caption{
    {Comparison of evolutionary strategies. }
    }
    \label{tab:tab_comparision_evo}
    \scalebox{.65}{
        \begin{tabular}{lccc}
            \toprule
            \multicolumn{1}{c}{Model} & Visual Evolve & Text Evolve & Capability-Aligned \\ \midrule
                MindGYM \citep{xu2025mindgym}               &    \xmark        &   \cmark    & -  \\
                R-CoT \citep{deng2024r}                     &    \cmark        &   \xmark    & -  \\
                MAVIS \citep{zhang2024mavis}                &    \cmark        &   \xmark    & -  \\
                OpenVLThinker \citep{deng2025openvlthinker} &    -             &   -         &  \xmark    \\
                \MethodName (Ours)                         &    \cmark        &   \cmark    &  \cmark      \\ \bottomrule
        \end{tabular}
    }\vspace{.25em}
\end{wraptable}

However, despite these promising developments, existing approaches face two key limitations:
(1) \textbf{Mismatched evolution of visual complexity and textual reasoning difficulty}. Prior methods often address visual and textual components in isolation. Some focus on generating visually complex scenes without meaningful reasoning tasks, while others emphasize textual complexity with simplistic visuals. 
This disconnect restricts the model’s ability to learn integrated cross-modal reasoning strategies.
(2) \textbf{The discrepancy between model capability and task difficulty}. As models improve over the course of training, their ability to tackle more complex tasks naturally increases. However, current approaches rely on static or manually defined difficulty schedules, which do not adapt to the model’s evolving capability. This misalignment can lead to inefficient training, either under-challenging the model or overwhelming it with excessively difficult data.

\begin{figure*}[t]
	\centering
        \includegraphics[width=0.99\textwidth]{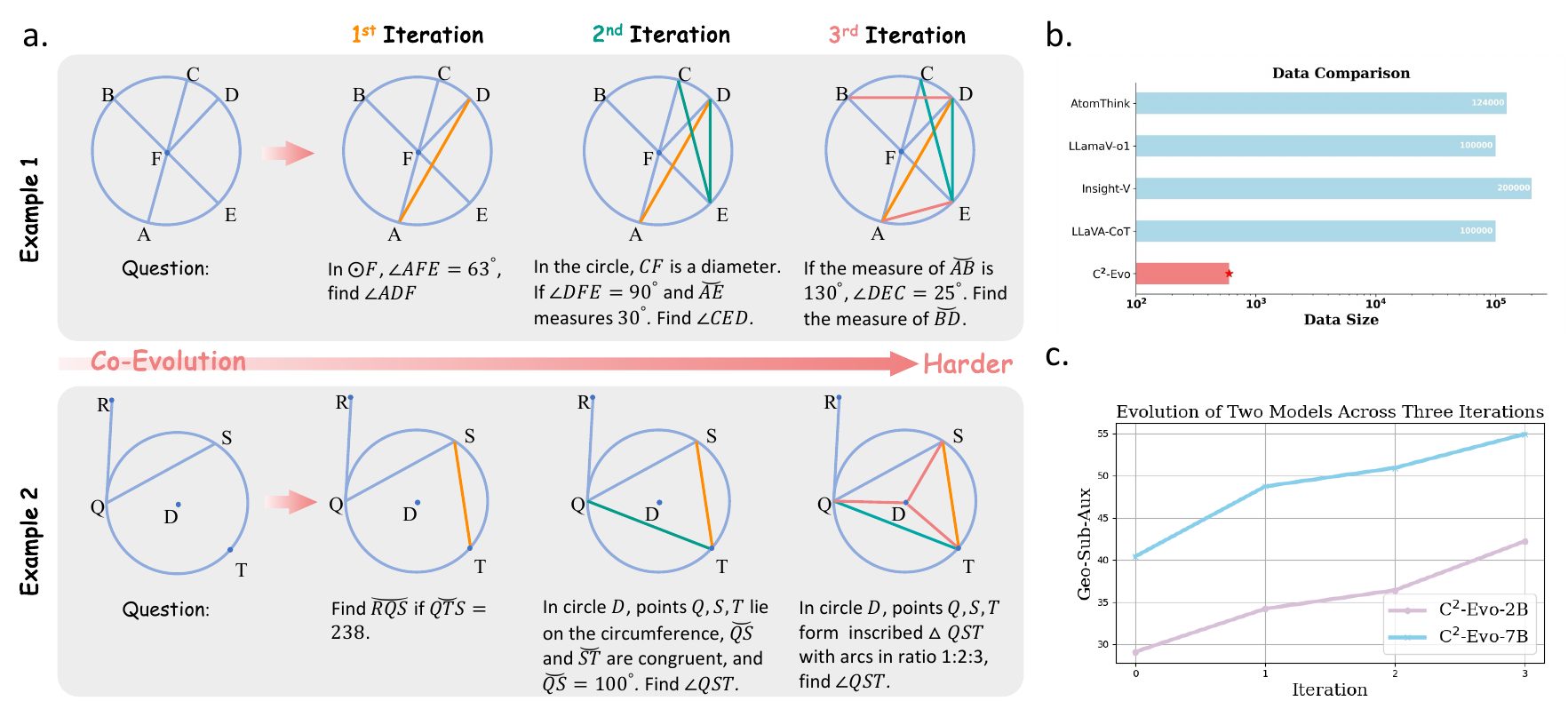}
        \caption{\textbf{a:} Co-evolution of visual-textual pairs with escalating task difficulty over three iterations.  
        \textbf{b:} Data usage comparison among different methods. Note that the x-axis (gate count) is on a log-scale.
        \textbf{c:} Performance of 2B and 7B models across three iterations.}
	\label{fig:fig_evoluation_example}
\end{figure*}

To address these challenges, we propose a fully automated, adaptive multimodal learning framework (\MethodName) that jointly evolves both the model and its training data in a closed-loop fashion, with a particular focus on geometric reasoning tasks. Table \ref{tab:tab_comparision_evo} summarizes the differences between our framework and state-of-the-art methods: Unlike existing methods that rely on static data, our method dynamically adjusts task complexity based on real-time assessments of model performance, ensuring a tighter coupling between model capability and data difficulty throughout the learning process.
{Specifically, to tackle \textbf{the challenge (1)}, 
we incorporate the process into a cross-modal data evolution loop, where
we integrate SKETCHPAD-generated complex geometric diagrams \cite{Hu2024VisualSS} with complex problem synthesis (\cf Figure \ref{fig:fig_evoluation_example} (a)).}
{GPT-4o is employed to generate auxiliary construction code, then executed via tools (\eg, Jupyter).}
Formal image descriptions are produced using the Doubao model. Sub-problems (4--10 per image) are then generated following predefined guidelines and composed into complex reasoning questions. These are filtered using GPT-4o to remove misaligned cases, and validated using DeepSeek's three-step reasoning framework to ensure consistency. This process yields a dataset with semantically aligned visual and textual components.
For \textbf{the challenge (2)}, we adopt a data-model evolution loop.
This loop utilizes data generated from the cross-modal data evolution loop and applies it to iteratively fine-tune the model using SFT and RL.
SFT maintains output structure and coherence, while RL improves generalization through rule-based optimization. 
We introduce a simple error-based filtering method that evaluates sample difficulty via prediction variance over $32$ generations.
{
By selecting samples with an error rate $ \geq 0.3$, which are prone to error yet still within the model's grasp and pose a meaningful challenge, the framework ensures that task difficulty remains aligned with model capability, achieving continuous improvement over iterations (\cf, Figure \ref{fig:fig_evoluation_example} (c)).
}
 
Finally, we investigate the impact of different data strategies and iterative training regimes on model performance. Our findings offer insights into the design of effective self-improving frameworks for improving complex multimodal reasoning and guiding the progressive evolution of vision-language models.

Our contributions are summarized as follows:

    $\bullet$ We propose a closed-loop self-improving framework, named {\MethodName}, that jointly evolves training data and model capabilities.

    $\bullet$ The proposed framework utilizes two co-evolution loops to improve the compatibility of cross-modal complexity and that between task difficulty and model capability.

    $\bullet$ Extensive experiments demonstrate the effectiveness of different data strategies and iterative training regimes, revealing their impact on self-improving frameworks.
\section{Related Work}
\label{sec:relat}

\noindent {\bf Multimodal Large Language Models.}  
Multimodal Large Language Models (MLLMs) have witnessed rapid advancements in recent years, resulting in significant breakthroughs in visual understanding and cross-modal reasoning within the field of artificial intelligence. 
Unlike traditional Vision-Language Models ~\cite{radford2021learning,jia2021scaling} (VLMs), which are typically trained from scratch on image-text pairs, MLLMs are generally built upon powerful pretrained text-only large language models (LLMs), and then further aligned with multimodal data, such as images and videos. 
Representative works such as BLIP-2 ~\cite{li2023blip}, LLaVA \cite{liu2023visual,liu2024improved,liu2024llavanext}, and MiniGPT-4 ~\cite{zhu2023minigpt} have demonstrated impressive zero-shot and few-shot generalization across various tasks, including image captioning, visual question answering, and image-text reasoning. LLaVA ~\cite{liu2023visual}, for example, pioneered the use of high-quality visual instruction data generated by GPT-4 ~\cite{achiam2023gpt} to fine-tune MLLMs, achieving significant success in visual dialogue and reasoning. This method has generated significant interest in research focused on creating multimodal datasets for tuning vision instructions ~\cite{chen2024sharegpt4v,liu2024improved}. 
BLIP-3 ~\cite{xue2024xgen} extends the single image input of BLIP-2 to interleaved multimodal data, using large-scale, high-quality, and diverse curated data with training recipes and post-training strategies to beat other contemporary competitors on various visual understanding tasks. 
Recently, several open-source MLLMs have significantly narrowed the performance gap with proprietary models like OpenAI's GPT-4o ~\cite{hurst2024gpt}. 
Among them, InternVL2 ~\cite{chen2024expanding}, Qwen2-VL ~\cite{bai2025qwen2}, SPHINX ~\cite{lin2023sphinx}, and MiniCPM-V ~\cite{yao2024minicpm} are particularly notable for their expanded application coverage, achieved through richer instruction-tuning datasets. 
In this study, we adopt a self-improving training paradigm to enhance the reasoning capabilities of MLLMs in complex scenarios. 

\noindent {\bf Reinforcement Learning.}  
Recent research has demonstrated that reinforcement learning (RL) can significantly enhance the reasoning capabilities of large language models (LLMs) such as OpenAI-o1 ~\cite{openai2024o1systemcard}. 
Some approaches have introduced RL-based mechanisms to facilitate test-time scaling, achieving notable success in tasks such as mathematical reasoning and code generation. 
Building on this progress, DeepSeek-R1 \cite{guo2025deepseek} proposed a rule-based reward strategy and adopted the Group Relative Policy Optimization (GRPO) ~\cite{shao2024deepseekmath} algorithm, demonstrating strong performance with only a few update steps. 
Motivated by the success of LLMs, recent studies  \citep{yang2025r1,huang2025vision,zhou2025r1,deng2025boosting,peng2025lmm,meng2025mm,zhang2025r1,peng2025skywork} have begun to explore the reasoning capabilities of MLLMs. 
For example, R1-OneVision ~\cite{yang2025r1} integrates supervised fine-tuning with RL to bridge the gap between visual perception and deep logical reasoning. 
Vision-R1 ~\cite{huang2025vision} generates cold-start initialization data and employs GRPO with hard format reward functions to enhance the emergent reasoning capabilities of MLLMs. VisualThinker-R1-Zero ~\cite{zhou2025r1} applies the R1 style to a base MLLM without supervised fine-tuning, surpassing traditional fine-tuning methods while exhibiting “visual aha moment” behaviors. 
LMM-R1 ~\cite{peng2025lmm} adopts a staged approach that begins with textual reasoning and advances toward complex multimodal reasoning tasks. MM-EUREKA~ \citep{meng2025mm} introduces a novel data filtering strategy that simultaneously removes both unsolvable and trivial cases, along with rejection samples, retaining only high-confidence instances.
The effectiveness of these methods in multimodal understanding tasks can be attributed to the presence of high-quality CoT datasets and the use of R1-style RL. Nonetheless, a significant drawback persists: the disconnect between the complexity of the training data and the difficulty of the tasks, particularly the inconsistency between the complexity of visual inputs and the challenges of textual reasoning. In this paper, we introduce a self-evolving training approach that alternates between RL and SFT. Our method dynamically modifies the multimodal training data to align with the model's reasoning abilities, ensuring that the complexities of both visual and textual elements are consistent throughout the training process. In contrast to MM-EUREKA, our method prioritizes samples that are both meaningfully challenging and conditionally accessible, striking a balance between task difficulty and model learnability. 



\noindent {\bf Self-Improvement.}  
Self-improvement  \citep{fernando2023promptbreeder,bhattarai2024heal,rosser2025agentbreeder} is a paradigm in which models generate and train on synthetic data generated from the same or other models. 
While self-improvement has been widely studied in the NLP domain, several works  \citep{zelikman2022star, gulcehre2023reinforced, singh2023beyond, lupidi2024source2synth,liang2024sheep,costello2025think} have explored the approach of first generating high-quality data and subsequently fine-tuning models on this data to achieve continuous performance improvement. 
For example,
STaR  \citep{zelikman2022star} introduces a bootstrapping mechanism that enhances LLM reasoning capabilities by iteratively generating and filtering "chain-of-thought" rationales, then fine-tuning the model on correct rationales to progressively improve performance.
ReST  \citep{gulcehre2023reinforced} integrates self-generated data with offline RL, alternating between a "Grow" phase that expands the dataset by generating multiple outputs per input, and an "Improve" phase that ranks and filters these outputs using a reward model based on human preferences.
Other works  \citep{lupidi2024source2synth} follow a similar approach of generating synthetic data, filtering out low-quality samples, and fine-tuning models on the filtered high-quality data.
Recent efforts  \citep{deng2024r, zhang2024mavis,trinh2024solving} have introduced synthetic data generation pipelines that leverage a visual data engine to produce visual images along with corresponding reasoning tasks. However, these methods primarily focus on expanding the training distribution. Our approach shifts the emphasis from passive data enrichment to self-improvement with a co-evolution mechanism.
A closely related work is OpenVLThinker  \citep{deng2025openvlthinker}, which introduces an iterative training paradigm in which models are exposed to progressively more complex tasks. However, their approach relies on manually defined task difficulty, which does not adapt to the evolving capabilities of the model.







\section{Method}
\label{sec:method}
In this section, we present the details of the proposed \MethodName~framework for self-improving reasoning. The key idea behind \MethodName is the joint evolution of multimodal data and models, thereby mitigating discrepancies not only between visual and textual modalities but also between task difficulty and model capabilities. We begin by formulating our task in Sec. \ref{sec: formulation}, and then introduce the two core components of \MethodName, namely the multimodal data co-evolution loop (Sec. \ref{sec: evo_data}), and the data and model co-evolution loop (Sec. \ref{sec: evo_data_and_model}), as shown in Figure \ref{fig:pipeline}.

\subsection{Task Formulation} \label{sec: formulation}




In this paper, we are given a pre-trained multimodal large language model (MLLM) denoted as $\pi_{\theta}$ and a dataset $\mathcal{D}$ consisting of image-question-answer triplets, namely, $\mathcal{D} = \{D_n=(I^n, Q^n, G^n)|n=1,...,|\mathcal{D}|\}$, where $I^n$, $Q^n$, and $G^n$ denote an image, questions related to $I^n$, and the corresponding answers of $Q^n$, respectively. Our goal is to improve the complex reasoning capability of $\pi_{\theta}$ by optimizing its parameters $\theta$ on $\mathcal{D}$ within $T$ iterations. Following recent studies ~\cite{deng2024r, zhang2024mavis, Hu2024VisualSS}, we assume that the necessary external tools and models are available, such as an oracle model that can generate and execute Python code to edit images ~\cite{Hu2024VisualSS}.

Specifically, let $t = 1,..., T$ denote the iteration step. At the beginning of the $t$-th iteration, we first conduct the multimodal data co-evolution loop to augment $\mathcal{D}_t$. This is achieved by prompting an MLLM \cite{hurst2024gpt} as the oracle model for solving each question in $\mathcal{D}_t$, generating step-by-step reasoning trajectories including necessary image augmentations like drawing auxiliary lines. For an arbitrary triplet $D^n_t \in \mathcal{D}_t$, the MLLM produces an action sequence $Ac_t^n$, corresponding Python code segment $C^n_t$, and a natural language reasoning trace $R^n_t$. The generated code $C^n_t$ is then executed using an external image editing tool \cite{Jupyter} to yield a more complex image $I^n_{t+1}$ that includes auxiliary augmentations. To ensure that the questions align with  $I^n_{t+1}$ with increased complexity, more challenging questions $Q^n_{t+1}$ and answers $G^n_{t+1}$ regarding $I^n_{t+1}$ are generated and curated, consequently a new multimodal triplet $D^n_{t+1}=(I^n_{t+1}, Q^n_{t+1}, G^n_{t+1})$ is obtained. We assess the difficulty of $D^n_{t+1}$ and if it is sufficiently challenging, we incorporate it into $\mathcal{D}_t$. Afterwards, we utilize the updated dataset $\mathcal{D}_{t+1}$ to train the base model $\pi_{\theta}$ through a combination of SFT and RL, where SFT establishes the initial reasoning structure and RL improves its generalization capability.

\subsection{Co-Evolving Multimodal Data} \label{sec: evo_data}
\begin{figure*}[t]
	\centering
        \includegraphics[width=0.99\textwidth]{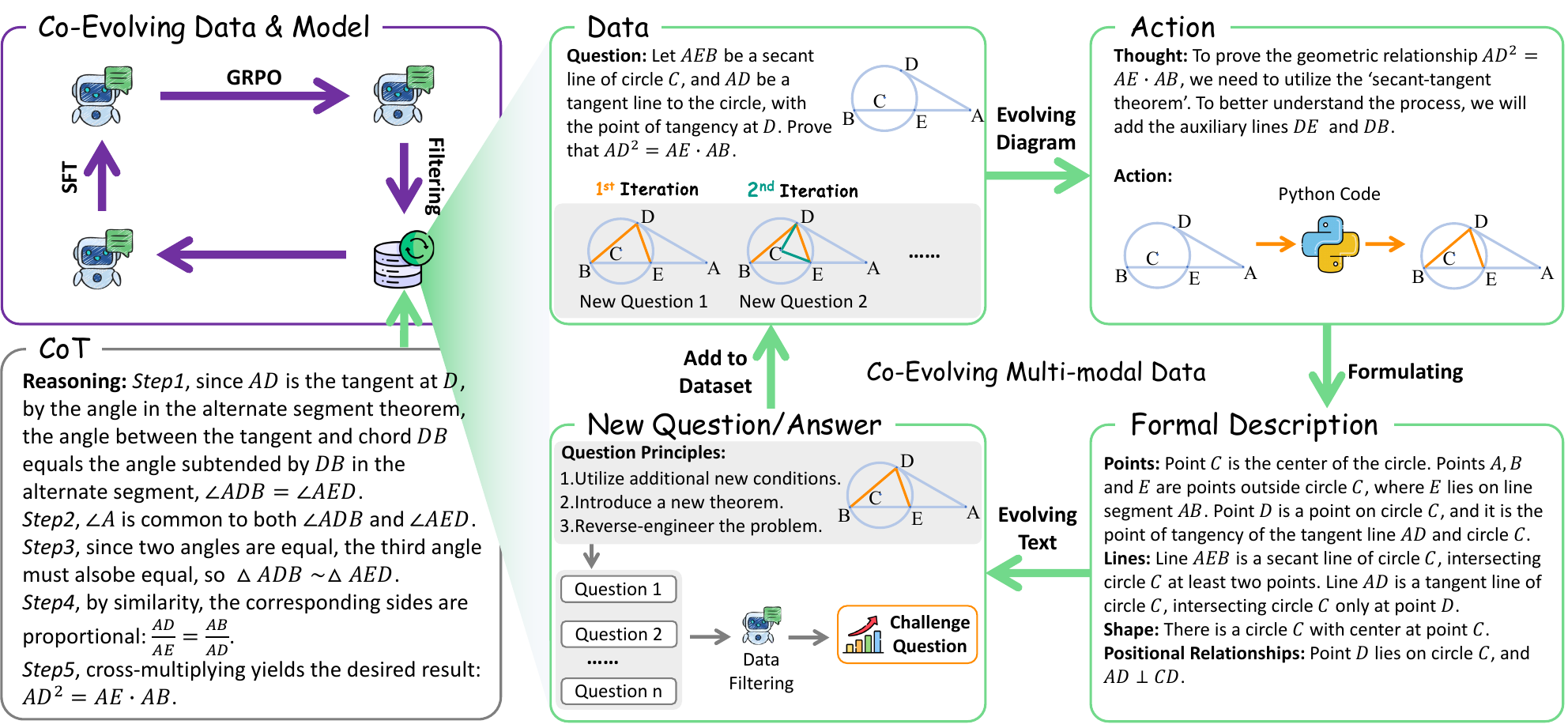}
        \caption{
        \textbf{The proposed {\MethodName} framework}. Starting from a base dataset and a base model, a co-evolving multimodal data loop iteratively synthesizes complex paired visual and textual samples. These samples are subsequently filtered and utilized in the co-evolving data $\&$ model loop to iteratively improve the reasoning performance of the model.
        }
	\label{fig:pipeline}
\end{figure*}
\begin{algorithm*}[t]
   \caption{{\MethodName} Algorithm}
   \label{alg:training_process}
    \textbf{Input:} Seed Dataset $\mathcal{D}$, Base Model $\pi_{\theta}$, Principles $P$, Number of Iterations $T$
     
    \textbf{Output:} Improved Model $\pi_{\theta, \text{RL}}^{T+1}$, Evolved Datasets $D_{T+1}$
     
   \begin{algorithmic}[1]
    \State Initialize $\pi_{\theta}^1 = \pi_{\theta}$ 

    \For{$t=1$ to $T$}
        \State \textcolor{gray}{$\triangleright$  \textbf{Co-Evolving Multimodal Data}} \Comment{see §\ref{sec: evo_data}}
        \For{$n=1$ to $|\mathcal{D}_t|$}
        \State $Ac_t^n, C_t^n, R_t^n \leftarrow $ generate action and reasoning conditioned on $D_t$
        \State $I_{t+1} \leftarrow$ apply auxiliary augmentations on $I_t$ by executing $(Ac_t^n, C_t)$
        \State $Q_{t+1}, A_{t+1} \leftarrow$ generate challenging questions and answers conditioned on $I_{t+1}$ and $P$
        \State $D_{t+1} \leftarrow$ add filtered tuple $(I_{t+1}, Q_{t+1}, A_{t+1})$ 
        \EndFor

        \State \textcolor{gray}{$\triangleright$  \textbf{Co-Evolving Data and Model}} \Comment{see §\ref{sec: evo_data_and_model}}
        \State {$D_{t}^{\text{train}} \leftarrow$ select data from $D_{t+1}$ using error rate evaluated with $\pi_{\theta, \text{RL}}^{t}$}     
        \State $\pi_{\theta, \text{SFT}}^{t+1} \leftarrow$ update $\pi_{\theta}^{t}$ with SFT on $D_t^{\text{train}}$ \Comment{using Equation \ref{eq_sft}}
        \State $\pi_{\theta, \text{RL}}^{t+1} \leftarrow$ update $\pi_{\theta, \text{SFT}}^{t+1}$ with GRPO on $D_t^{\text{train}}$ \Comment{using Equation \ref{eq_rl}}
    \EndFor

   \end{algorithmic}

    
    


\end{algorithm*}
The co-evolving multimodal data loop is introduced to mitigate the complexity discrepancy between visual and textual data, which can be further divided into an action and reasoning generation process and a challenging question generation process.


\noindent \textbf{Action and Reasoning Generation.} 
Inspired by SKETCHPAD \citep{Hu2024VisualSS}, we introduce a two-stage framework designed to enable the precise construction of auxiliary lines in geometric diagrams.
In the first stage, we extract the coordinates of key points using Optical Character Recognition (OCR) and other vision-based techniques \footnote{\url{https://github.com/lupantech/InterGPS/blob/main/diagram_parser}}. This coordinate-level representation ensures robustness and consistency in subsequent diagram transformations, particularly under increased geometric complexity. 
Leveraging the extracted coordinates and image $I_t$ (we omit the superscript $n$ for conciseness), 
{we prompt\footnote{All detailed prompts are provided in the Supplementary Material.}} GPT-4o to determine whether auxiliary image augmentations are needed to facilitate problem solving. 
This generates a decision, denoted as \textit{Thought}, indicating whether auxiliary augmentations are necessary and specifying their types (\eg, parallel lines, perpendicular lines, and connecting lines). Subsequently, it produces the updated image $I_{t+1}$ with newly added auxiliary augmentations, as illustrated in Figure \ref{fig:pipeline}. In addition to $I_{t+1}$, the pair $(I_t, Q_t)$ is also fed to GPT-4o to simultaneously generate the full reasoning trace $R_t$, which is used in the subsequent generation of challenging questions.

\noindent \textbf{Challenging Question Generation.}
This step aims to match the level of problem difficulty with the degree of image complexity.
Alternative to GPT-4o, we utilize the Doubao model \cite{Doubao-1.5-pro} to generate a formatted description $F_t$ conditioned on the generated images $I_{t+1}$, as in practice we find it provides more precise descriptions.
To promote diversity in the generated sub-problems, we define a set of guiding principles $P$:

1) \emph{Geometric Constraints}: We extract geometric constraints (\eg, perpendicularity, equality, and sub-images) from the formal description $F_t$.

2) \emph{New Theorems and Concepts}: We incorporate relevant mathematical theorems and conceptual principles (\eg, the properties of a right triangle imply the Pythagorean theorem) that are closely related to the formal description $F_t$.

3) \emph{Backward Reasoning}: We also include the concept (\eg, the Tangent-Secant theorem) of sub-steps in the inference trace (\ie, $R_t$).

Using these principles, we prompt DeepSeek-R1 with the tuples $(F_t, R_t, P)$ to generate a diverse set of sub-problems ($q_1, q_2, \ldots, q_m$, where $m$ typically ranges from $4$ to $10$, depending on the complexity of the image).
To mitigate the inherent limitations of formal language in capturing visual content, we incorporate the role-playing strategy from R1-Onevision  \citep{Yang2025R1OnevisionAG}.
By analyzing the differences and commonalities among sub-problems ($q_1, q_2, \ldots, q_m$), we align them with corresponding sub-image elements to compose challenging geometric reasoning questions.
Specifically, by combining the sub-problems ($q_1, q_2, \ldots, q_m$) with their associated sub-image descriptions $F_t$, we prompt DeepSeek-R1 to compose challenging questions.
For example, given two sub-problems $q_1$ and $q_2$ that address different parts of an image, we combine them into a more challenging question $Q_{t+1}$ by leveraging their thematic and contextual connections. 


\begin{wrapfigure}[16]{r}{.62\linewidth}
    \centering
    \vspace*{-1.7em}
    \includegraphics[width=\linewidth]{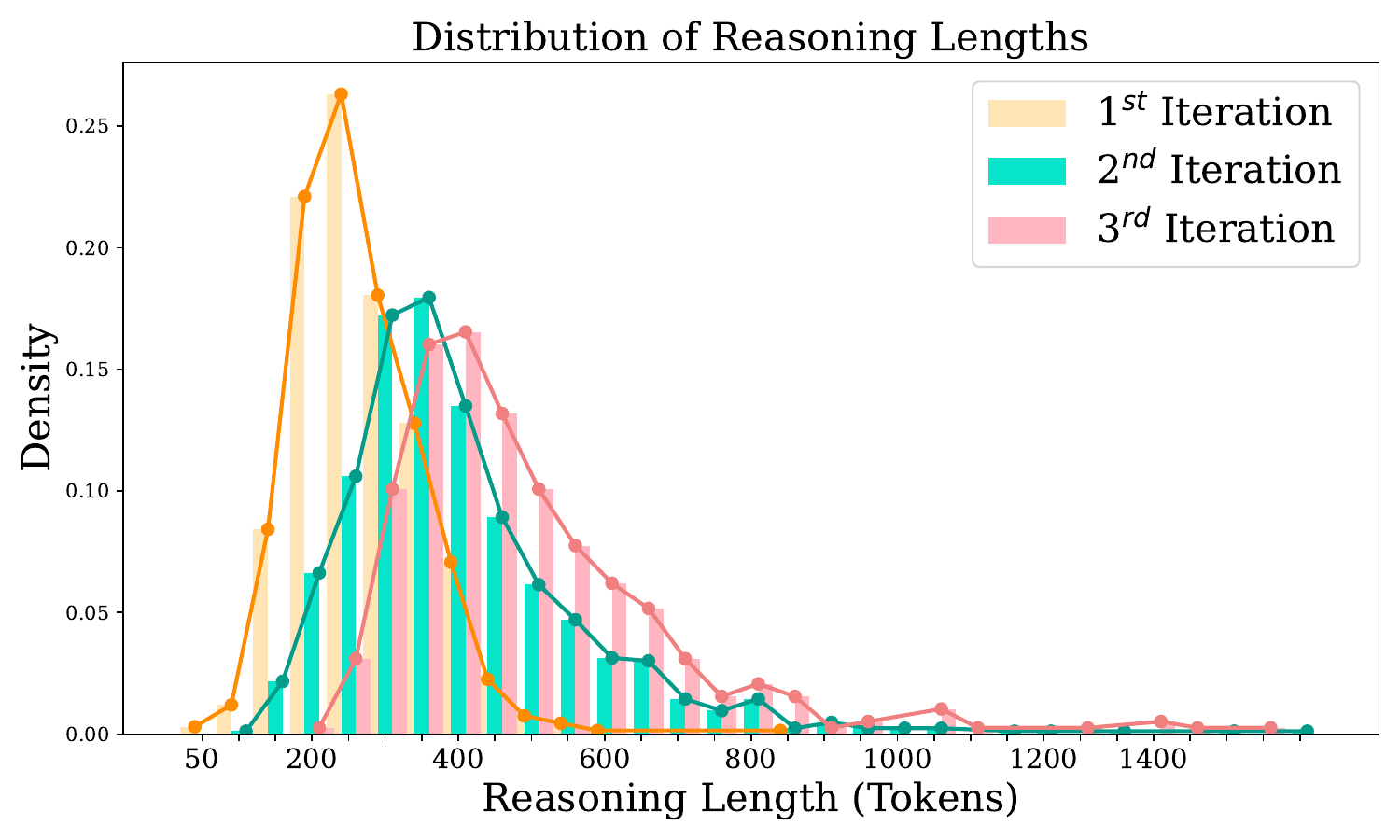}
    \caption{
        \textbf{Evolution of reasoning length} (tokens) over three iterations.
    }
    \label{fig:fig_reasoning_length}
\end{wrapfigure}%
To obtain corresponding answers, we feed each generated question $Q_{t+1}$ together with its formal description $F_t$ into DeepSeek-R1 three times. For each input, the model produces a step-by-step reasoning trace $R_{t+1}$ and the final answer $A_{t+1}$. 
We then retain only those samples with three answers $A_{t+1, 1}, A_{t+1, 2}, A_{t+1, 3}$ are equal.
We further prompt GPT-4o to filter out any questions that are inconsistent with the original image constraints, ensuring alignment between the question and the visual content. 
As illustrated in Figure \ref{fig:fig_reasoning_length}, the increasing length of the reasoning trajectories serves as a reliable indicator of problem difficulty, aligning with our data evolution design.

\subsection{Co-Evolving Data and Model} \label{sec: evo_data_and_model}
\noindent \textbf{SFT to Follow Reasoning Structure}
This stage aims to unify the model's reasoning format with the structure needed during the later reinforcement learning (RL) phase, ensuring compatibility through a standardized template (\ie, <think></think><answer></answer>).
The model is trained on the previously constructed triplets $(I_{t+1}, Q_t, R_t)$, learning to generate the reasoning trace and final answer in a structured format.
The corresponding training objective is formulated as follows:
\begin{equation}
    \mathcal{L}_\text{SFT} = -\mathbb{E}_{(I,Q,R) \sim D}[\text{log}(\pi_{\theta}(R |I,Q)].
\label{eq_sft}
\end{equation}

\noindent \textbf{Group Relative Policy Optimization}
Based on the above obtained $\pi_{\theta,\text{SFT}}$ model, we employ two reward rules in conjunction with the GRPO algorithm to further refine the policy model. These rewards are designed as follows:

1) Accuracy Reward: This reward evaluates the correctness of the final answer $A$ by processing the final answer via regular expressions and verifying it against the ground truth $G$.

2) Format Reward: To ensure consistent and well-structured reasoning trajectories, we define a reward based on the model's adherence to the predefined format <think> $\cdots$ </think>. 

Additionally, we enforce sequential consistency within the reasoning trace by requiring that each step be explicitly arranged in the correct order. Responses that violate this ordering are penalized during training. The training objective with GRPO is defined as,
\begin{equation}
\begin{split}
    \mathcal{L}_{\text{RL}} = \mathbb{E} \left[ \min \left( r_t(\theta) A_t, \text{clip}(r_t(\theta), 1-\varepsilon, 1+\varepsilon) A_t \right) - \beta D_{\text{KL}}[\pi_{\theta} \| \pi_{\text{old}}] \right]
\end{split}.
\label{eq_rl}
\end{equation}

\noindent \textbf{Iterative Refinements and Filtering}

\begin{wrapfigure}[11]{r}{.34\linewidth}
    \centering
    \vspace*{-3.1em}
    \includegraphics[width=\linewidth]{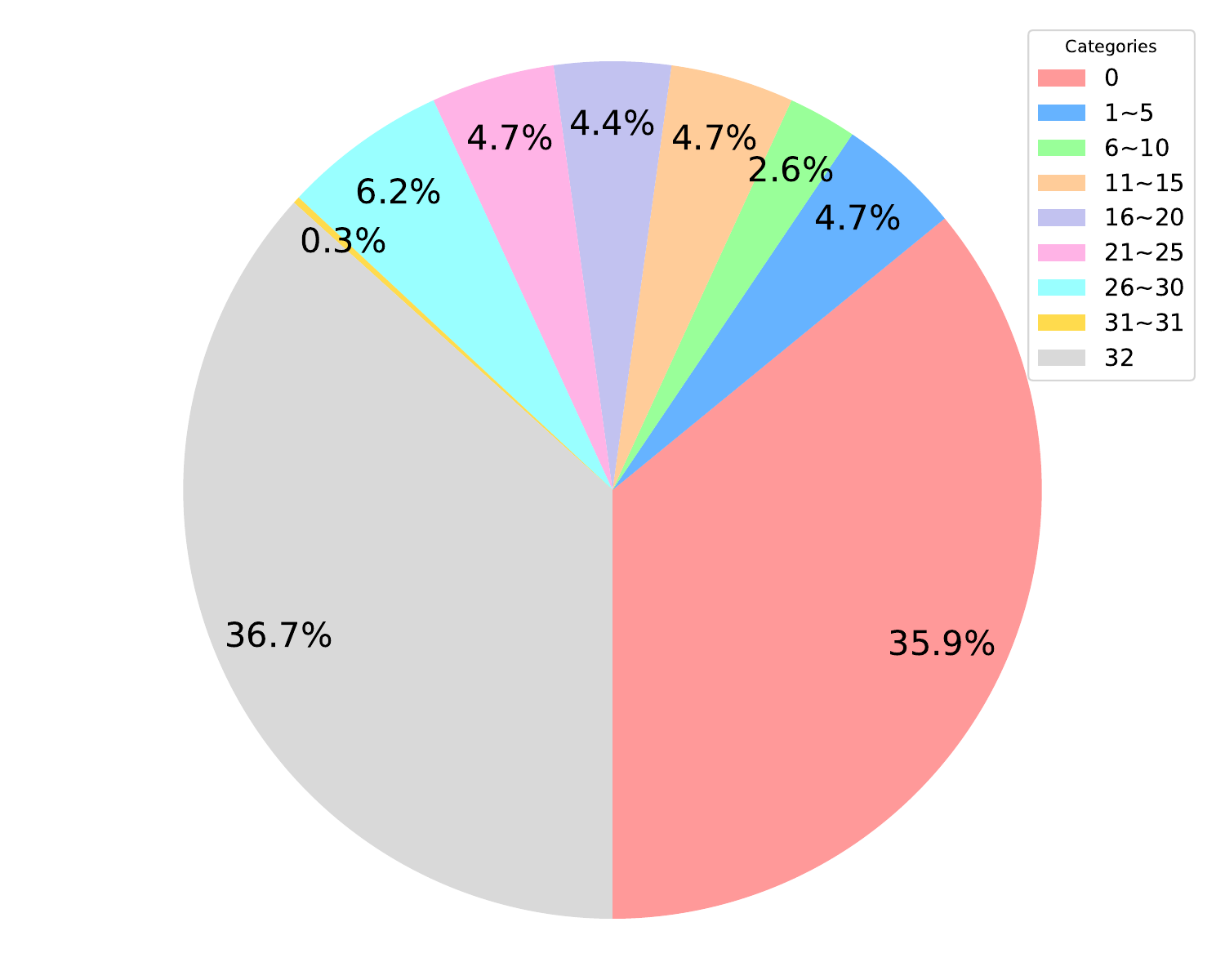}
    \caption{
    Assessment of third iteration training data difficulty using second iteration model $\pi_{\theta, \text{RL}}^2$.
    }
    \label{fig:fig_eri}
\end{wrapfigure}%
In the $t$-th iteration, we perform SFT followed by RL training using all post-filtered data, resulting in an updated model $\pi_{\theta, RL}^{t+1}$ and a new dataset $\mathcal{D}_{t+1}$ for the next round.
To align the difficulty of $\mathcal{D}$ with the current capability of $\pi_{\theta}$, we forward each sample through the model $32$ times and compute the error rate as the proportion of times the model generated a wrong answer.
As illustrated in Figure \ref{fig:fig_eri}, we evaluate the difficulty of the data for each model obtained from RL training.
We then retain only the samples with an error rate greater than $0.3$ to form the training set for the next iteration, ensuring that the data complexity remains aligned with model capability.
\section{Experiment}
\label{sec:experiment}

\subsection{Benchmark} \label{sec:5_1}

\noindent \textbf{Dataset}.
We conduct experiments on the Geometry3k  \citep{Lu2021InterGPSIG} dataset and use it as our base dataset for multimodal data evolution. Geometry3k is a mathematical benchmark dataset that comprises 3,002 geometry problems, divided into 2,101 training examples, 300 for validation, and 601 for testing. Each problem is accompanied by a corresponding geometric diagram, a natural language description, and formal language annotations.

\noindent \textbf{Implementation Details}.
In our experiments, we adopt two state-of-the-art open-source MLLMs, \ie, Qwen2-VL-2B \citep {wang2024qwen2} and Qwen2-VL-7B \citep {wang2024qwen2}.
For the policy warm-up phase, we employ the LLaMA-Factory\footnote{\url{https://github.com/hiyouga/LLaMA-Factory}.} framework with a batch size of $128$ and a learning rate of $1e-5$.
For the RL phase, we use the VLM-R1\footnote{\url{https://github.com/om-ai-lab/VLM-R1}.} framework. In the first iteration, we perform $32$ rollouts per question, reducing to $8$ in subsequent iterations.
The temperature is set to the default value of $0.9$, and the KL divergence coefficient $\beta$ in Equation \ref{eq_rl} is set to $0$.
All experiments are conducted on $32$ NVIDIA V100-$32$GB GPUs.

We evaluate {\MethodName} on several multimodal reasoning benchmarks, including {Geo-Sub} from Geometry3k-test 
 \citep{Lu2021InterGPSIG}, MathVista \citep{lu2023mathvista}, and MathVerse \citep{zhang2024mathverse}.
To provide a more comprehensive evaluation of model performance, we construct a specialized test subset by sampling images from Geometry3K-test that require the use of auxiliary lines to solve. This results in a focused benchmark containing  $274$ images.
When evaluated on the original images, we denote the setting as {Geo-Sub}. Using the corresponding images with added auxiliary lines, we refer to the setting as {Geo-Sub-Aux}. 
Further details can be found in the Supplementary Material.

\subsection{Main Results} \label{sec:5_2}

\begin{table}[]
\centering
\caption{Main experimental results. 
For MathVista benchmark, we have specifically compared all models on three sub-tasks that are highly related to mathematical reasoning: geometry reasoning (GEO), algebraic reasoning (ARI) and geometry problem solving (GPS). The result ($^\dag$) is collected from original papers, R1-VL\citep{zhang2025r1} and R1-Onevison~\citep{Yang2025R1OnevisionAG}. The remaining results are reproduced under the same experimental setting.
}
\label{tab:tab_main_results}
\scalebox{0.9}{
\begin{tabular}{lccccccc}
\toprule
Methods                      & Data Amount        & Geo-Sub & \multicolumn{1}{l}{Geo-Sub-Aux} & \multicolumn{4}{c}{MathVista}              \\ \cline{5-8}
                             &                    &         & \multicolumn{1}{l}{}            & GEO   & ARI   & GPS   & ALL                          \\ \midrule
\textit{Closed-Source Model} &                    &         & \multicolumn{1}{l}{}            &       &       &       &                              \\
GPT-4o \citep{hurst2024gpt}                       &                    & -       & -                               & -     & -     & -     & 63.8$^\dag$                   \\ \midrule
\textit{Reasoning Model}     &                    &         & -                               &       &       &       &                              \\
LLamaV-o1-11B\citep{thawakar2025llamav}                & \textgreater{}100k & -       & -                               & -     & -     & -     & 54.4$^\dag$                      \\
Insight-V-8B\citep{dong2024insight}                 & 200K               & -       & -                               & -     & -     & -     & 49.8$^\dag$                     \\
MGT-PerceReason \citep{Peng2025LMMR1E3}                & 65k & -       & -                               & -     & -     & -     & 63.2$^\dag$                   \\
R1-Onevision-7B \citep{Yang2025R1OnevisionAG}             & 10k               & -       & -                               & -     & -     & -     & 64.1$^\dag$                    \\ 
R1-VL-7B \citep{zhang2025r1}                  & 10k (SFT 120k)                  & -     & - & - & - & -                           & 63.6$^\dag$                          \\
\midrule
Qwen2-VL-2B \citep{wang2024qwen2}                  & -                  & 28.0    & 29.1                            & -     & -     & -     & 43.0$^\dag$                   \\
\MethodName-$1^{st}$                        & 0.6k               & 32.0    & 34.2                            & 38.0  & 40.0  & 38.0  & 49.1                      \\
\MethodName-$2^{nd}$                        & 0.6k               & 35.6    & 36.4                            & 38.0  & 38.0  & 38.0  & 49.3                 \\
\rowcolor{mycustomcolor}  \MethodName-$3^{rd}$                        & 0.4k               & 38.2    & 42.2                            & 40.0  & 40.0  & 38.0  & 50.2                       \\ \midrule
Qwen2-VL-7B \citep{wang2024qwen2}                  & -                  & 40.4    & 40.4                            & 50.0  & 50.0  & 49.0  & 60.0                    \\
\MethodName-$1^{st}$                        & 0.6k               & 45.5    & 46.9                            & 55.0  & 57.0  & 54.0  & 62.1  \\
\MethodName-$2^{nd}$                         & 0.6k               & 46.9    & 48.0                            & 56.0  & 58.0  & 56.0  & 62.4                      \\
\rowcolor{mycustomcolor}  \MethodName-$3^{rd}$                      & 0.4k               & 50.9    & 52.4                            & 59.0  & 59.0  & 60.0  & 63.2                      \\ \bottomrule
\end{tabular}}
\end{table}

As shown in Table \ref{tab:tab_main_results}, {\MethodName} demonstrates a significant improvement over Qwen2-VL-7B and Qwen2-VL-2B across three benchmarks through three iterations, with geometric reasoning accuracy increasing notably from $40.4$ to $54.9$.
The progressive improvement observed across three iterations highlights the effectiveness of our self-improving strategy in enhancing reasoning capabilities.
Furthermore, compared to prior reasoning models, our approach achieves superior performance while leveraging less than $1\%$ of the training data, with performance approaching that of GPT-4o.


\begin{wrapfigure}[15]{r}{.7\linewidth}
    \centering
    \vspace*{-1em}
    \includegraphics[width=\linewidth]{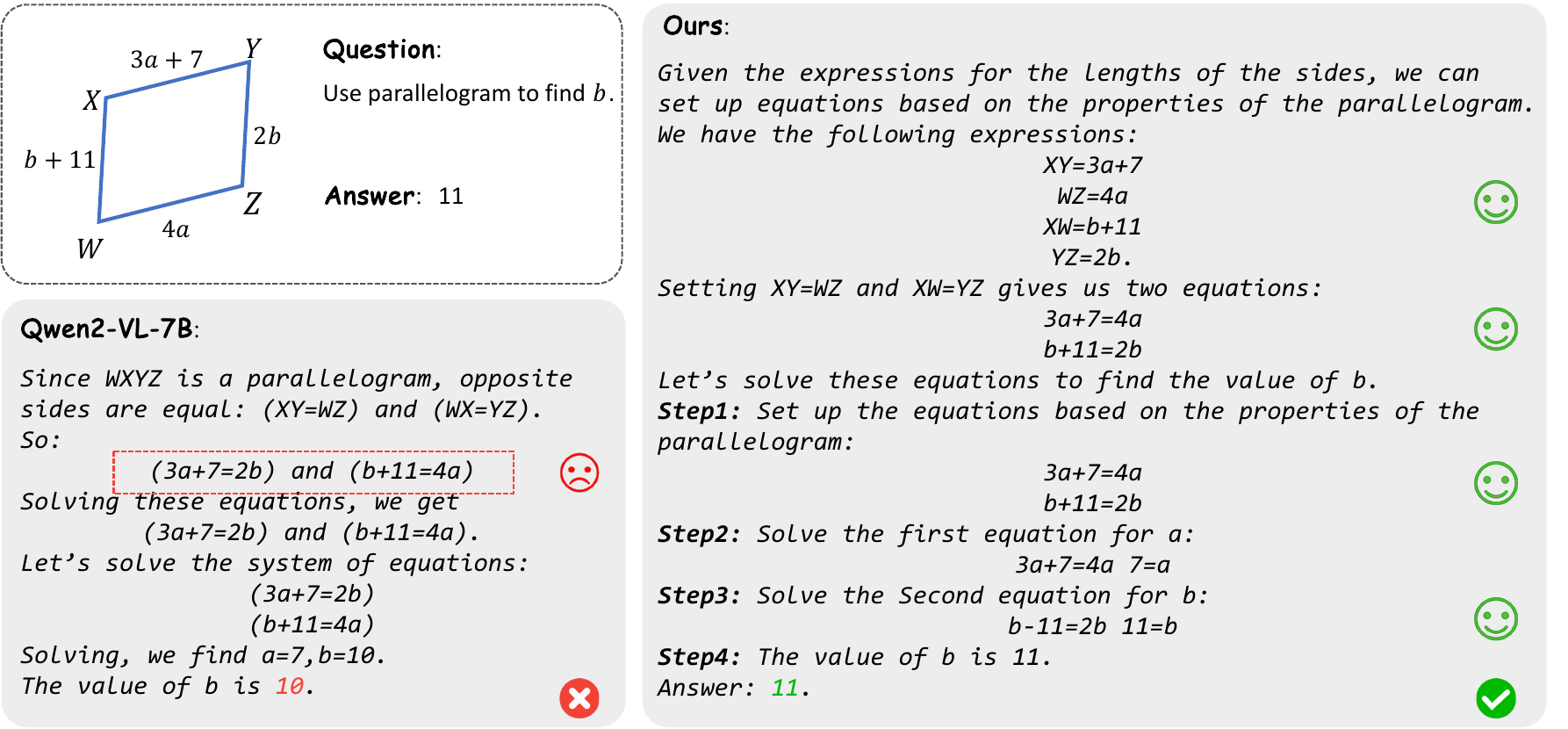}
    \caption{
        Qualitative comparison of geometric reasoning.
    }
    \label{fig:figexample}
\end{wrapfigure}%

Figure \ref{fig:figexample} illustrates the model’s behavior on a geometry problem involving a parallelogram. The response generated by Qwen2-VL-7B is relatively short but lacks a thorough reasoning process. It contains inaccuracies in both formula application and computational steps, leading to an incorrect final answer.
In contrast, our {\MethodName} produces a well-structured and logically coherent solution. It begins with a thorough analysis of the given conditions, followed by a step-by-step deductive process that ensures correctness at each stage, ultimately yielding the accurate final result.

\begin{table*}[h]
\begin{minipage}{0.475\textwidth} 
\caption{Comparison of alternating vs. consecutive RL training schedules with progressive model updates.}
\vspace{-10pt}
\label{tab:tabSFTRL1}
\begin{center}
\resizebox{\linewidth}{!}{
\begin{tabular}{llccc}
\toprule
Iteration               & \multicolumn{1}{c}{Model} & \multicolumn{1}{l}{SFT} & \multicolumn{1}{l}{RL} & Geo-Sub-Aux \\ \midrule
     \multirow{2}{*}{Second}                    & $\pi_{\theta}^1$          & \cmark         & \cmark              & 40.7       \\
                        & $\pi_{\theta}^1$          & -              & \cmark              & \textbf{50.6}       \\ \midrule
        \multirow{2}{*}{Third}                  & $\pi_{\theta}^2$          & \cmark         & \cmark              & 42.9           \\
                        & $\pi_{\theta}^2$          & -              & \cmark              & \textbf{54.9}           \\ \bottomrule
\end{tabular}
}
\label{fig:hadamard}
\end{center}
\end{minipage}
\begin{minipage}{0.475\textwidth}
\caption{Comparison of alternating vs. consecutive RL training schedules from the initial model.}
\vspace{-10pt}
\label{tab:tab_initial}
\begin{center}
\resizebox{\linewidth}{!}{
\begin{tabular}{llccc}
\toprule
Iteration               & \multicolumn{1}{c}{Model} & \multicolumn{1}{l}{SFT} & \multicolumn{1}{l}{RL} & Geo-Sub-Aux \\ \midrule
        \multirow{2}{*}{Second}                  & $\pi_{\theta}^0$          & \cmark         & \cmark              & \textbf{48.4}       \\
                        & $\pi_{\theta}^0$          & -              & \cmark              & 44.0       \\ \midrule
        \multirow{2}{*}{Third}                  & $\pi_{\theta}^0$          & \cmark         & \cmark              & \textbf{46.7}           \\
                        & $\pi_{\theta}^0$          & -              & \cmark              & 44.4           \\ \bottomrule
\end{tabular}
}
\end{center}
\end{minipage}
\end{table*}
\noindent \textbf{The influence of different iteration strategies.}
Table \ref{tab:tabSFTRL1} presents the results of different iteration strategies. Fine-tuning $\pi_{\theta}^1$ with additional SFT in the second iteration leads to performance degradation. In contrast, further refining the model through reinforcement learning yields improved results. 
As shown in Table \ref{tab:tab_initial}, when the second iteration model is trained via SFT and RL starting from the initial model $\pi_{\theta}$, the same performance degradation is not observed.
However, the resulting model underperforms compared to when the training is warm-started from $\pi_{\theta}^1$.
Similar trends are observed in the third iteration.

\begin{wraptable}[11]{r}{.51\linewidth}
    \centering\setlength{\tabcolsep}{4.5pt}
    \vspace{-9pt}
    \caption{\label{tab:tab_error_rate}%
        Comparison of error-rates based on the second iteration.
    } %
    \scalebox{.85}{
    \begin{tabular}{lcc}
        \toprule
        \multicolumn{1}{c}{Models} & \multicolumn{1}{l}{Data Size ($D_t/D_{1 \sim t}$)} & Geo-Sub-Aux \\ \midrule
        Qwen2-VL-7B                & -                             &  38.2/40.4                    \\ \midrule
        Fullset                    & 0.83K/1.48K                   & 48.4/46.6           \\
        $\pi_{\theta}^1$- 0.3                   & 0.48K/0.64K                   & 49.1/\textbf{50.9}         \\
        $\pi_{\theta}^1$- 0.6                   & 0.44K/0.54K                   & \textbf{50.9}/47.3          \\
        $\pi_{\theta}^1$- 0.9                   & 0.38K/0.44K                   & 48.4/47.3            \\
        $\pi_{\theta}^1$- 1.0                   & 0.34K/0.35K                   & 46.2/49.5           \\ \bottomrule
\end{tabular}
    }
\end{wraptable}
\noindent \textbf{The Influence of error-rate.}
As described in Section \ref{sec: evo_data_and_model}, we conduct an extensive parameter analysis over varying values of error-rate, which governs the complexity of data within each iteration.
Table \ref{tab:tab_error_rate} shows the performance of models trained solely on data from the current iteration $D_t$, compared to those that also incorporate historical data $D_{1, \ldots, t-1}$.
Utilizing the complete dataset does not necessarily optimize the model's reasoning capabilities.
Specifically, as the complexity of the training data increases, relying exclusively on error-prone samples results in a decline performance.
Based on this analysis, error-rate$ \geq 0.3$ is adopted to as the basis for our final setting.



\begin{wraptable}[7]{r}{.4\linewidth}
    \centering\setlength{\tabcolsep}{4.5pt}
    \vspace{-1.3em}
    \caption{\label{tab:tab_original_data}%
        Training comparison between original and complex image datasets.
    } %
    \scalebox{.85}{
    \begin{tabular}{lcc}
        \toprule
        \multicolumn{1}{c}{Model} & \multicolumn{1}{l}{Original Data} & Complex Data \\ \midrule
        \MethodName-$1^{st}$                          & 47.27                             & 48.7         \\
        \MethodName-$2^{nd}$                         & 45.82                             & 50.9         \\
        \MethodName-$3^{rd}$                          & 51.27                             & 54.9         \\ \bottomrule
        \end{tabular}
    } 
\end{wraptable}
\noindent \textbf{The Influence of training data.}
Table \ref{tab:tab_original_data} presents a comparison of the 7B model trained on complexified data (complex images with complex text) versus the original data (original images with complex text).
The results show that jointly complexifying both images and text leads to better performance, highlighting the importance of image complexity in training.

\section{Conclusion}
\label{sec:conclusion}

In this paper, we propose a closed-loop self-improving framework (\MethodName), a multi-dimensional evolution framework that operates through two interleaved loops: \textbf{a cross-modal data evolution loop }and \textbf{a data-model co-evolution loop}.
Recent studies often suffer from the decoupling of textual and visual evolution, as well as a mismatch between model capability and task difficulty. To address these limitations, we introduce a synchronized co-evolution mechanism in which auxiliary guidance is employed to progressively increase the complexity of image data, while the resulting complex visuals are used to generate increasingly challenging reasoning tasks. This ensures the joint evolution of both modalities.
Furthermore, by leveraging the error-based filtering method, the model selects samples that align with the current model's blind spots or underdeveloped reasoning capabilities, thereby maintaining consistency between data complexity and model capability through iterative training. 
Extensive experiments demonstrate the effectiveness of our multi-iteration evolution training strategy. The different data curation strategies and iterative mechanisms not only improve performance but also offer promising directions for future research in self-improving methods.

\noindent \textbf{Limitations.}
While our approach uses self-iterative evolution to improve geometric reasoning and auxiliary line construction, it mainly relies on a dataset limited to mathematical reasoning problems. The scope of the data may not fully represent the diversity of real-world geometric problems.
Additionally, relying on synthetic data from self-iteration may bias the model toward easier-to-evolve solutions, potentially missing less common but valid approaches to geometric problems.

\noindent \textbf{Future Work.}
The present study is limited to the domain of geometric reasoning. Future work will explore the applicability of our framework to more diverse domains, such as mathematical function understanding and general vision-language tasks. We also intend to investigate the impact of model scale on performance, aiming to understand how architectural capacity interacts with evolutionary training dynamics.

{\small
\bibliographystyle{splncs04}
\bibliography{egbib}
}


\setcounter{section}{0}
\renewcommand\thefigure{\Roman{figure}}
\renewcommand\thetable{\Roman{table}}
\renewcommand\thesection{\Alph{section}}
\renewcommand\theequation{\roman{equation}}


\section{Experimental Supplement}
\label{sec:sec_data_details}

\subsection{Implementation Details} 
\begin{table}[ht]
\caption{\label{tab:tab_parameter}
    Training parameters of Qwen2-VL-2B/7B model.
    } %
\scalebox{0.9}{
\begin{tabular}{lcccc}
\toprule
Parater       & Qwen2-VL-2B-SFT & Qwen2-VL-2B-RL & Qwen2-VL-7B-SFT & Qwen2-VL-7B-RL \\ \midrule
Learning Rate & 1e-5            & 1e-6           & 1e-5            & 1e-6           \\
Epochs        & 2               & 2              & 2               & 2              \\
Batch Size    & 128             & 128            & 128             & 128            \\
Precision     & fp16            & fp16           & fp16            & fp16           \\
GPU           & 32 V100         & 32 V100        & 32 V100         & 32 V100        \\
Temperature   & -               & 0.9            & -               & 0.9            \\ \bottomrule
\end{tabular}}
\end{table}
In this section, we provide more implementation details for the \MethodName.
In Table \ref{tab:tab_parameter}, we provide the training parameters of Qwen2-VL-2B/7B during the SFT and RL stages to facilitate future work.

We evaluate {\MethodName} on three reasoning benchmarks, Geo-Sub from  Geometry3K-test \citep{Lu2021InterGPSIG}, MathVista \citep{lu2023mathvista}, and MathVerse \citep{zhang2024mathverse}. For MathVista, we use the Test Mini split (\cf, around 1,000 samples). For MathVerse, we use the full dataset.

\subsection{The difference between Geo-Sub and Geo-Sub-Aux.} 

\begin{wrapfigure}[10]{r}{.4\linewidth}
    \centering
    \vspace*{-2.5em}
    \includegraphics[width=\linewidth]{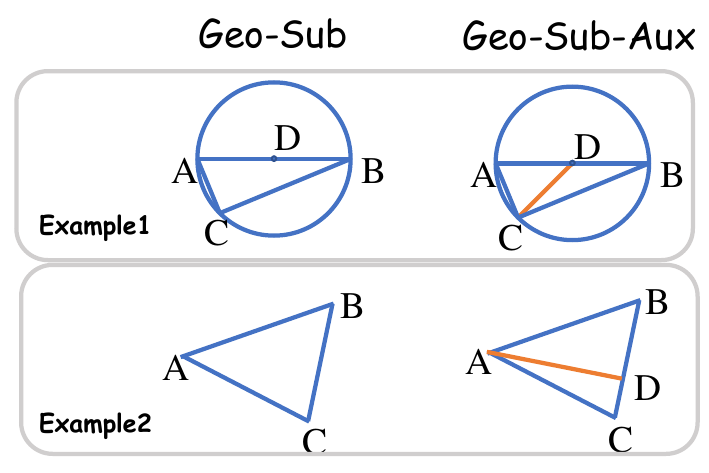}
    \caption{
        Data comparison between Geo-Sub and Geo-Sub-Aux.
    }
    \label{fig:fig_test_set}
\end{wrapfigure}%
We have introduced auxiliary lines into the Geo-Sub-Aux dataset to facilitate problem solving and improve model alignment during training. Originating from Geometry3k\citep{Lu2021InterGPSIG}, the Geo-Sub dataset comprises $275$ samples. The left image illustrates the structure of the Geo-Sub dataset  (\cf, Figure \ref{fig:fig_test_set} left), whereas the right image showcases the Geo-Sub-Aux dataset  (\cf, Figure \ref{fig:fig_test_set} right), highlighted by its additional auxiliary lines.

\subsection{ Data Difficulties Across Three Iterations.} 

\begin{figure*}[htbp]
    \centering
    \begin{minipage}[b]{0.32\textwidth}
        \includegraphics[width=\linewidth]{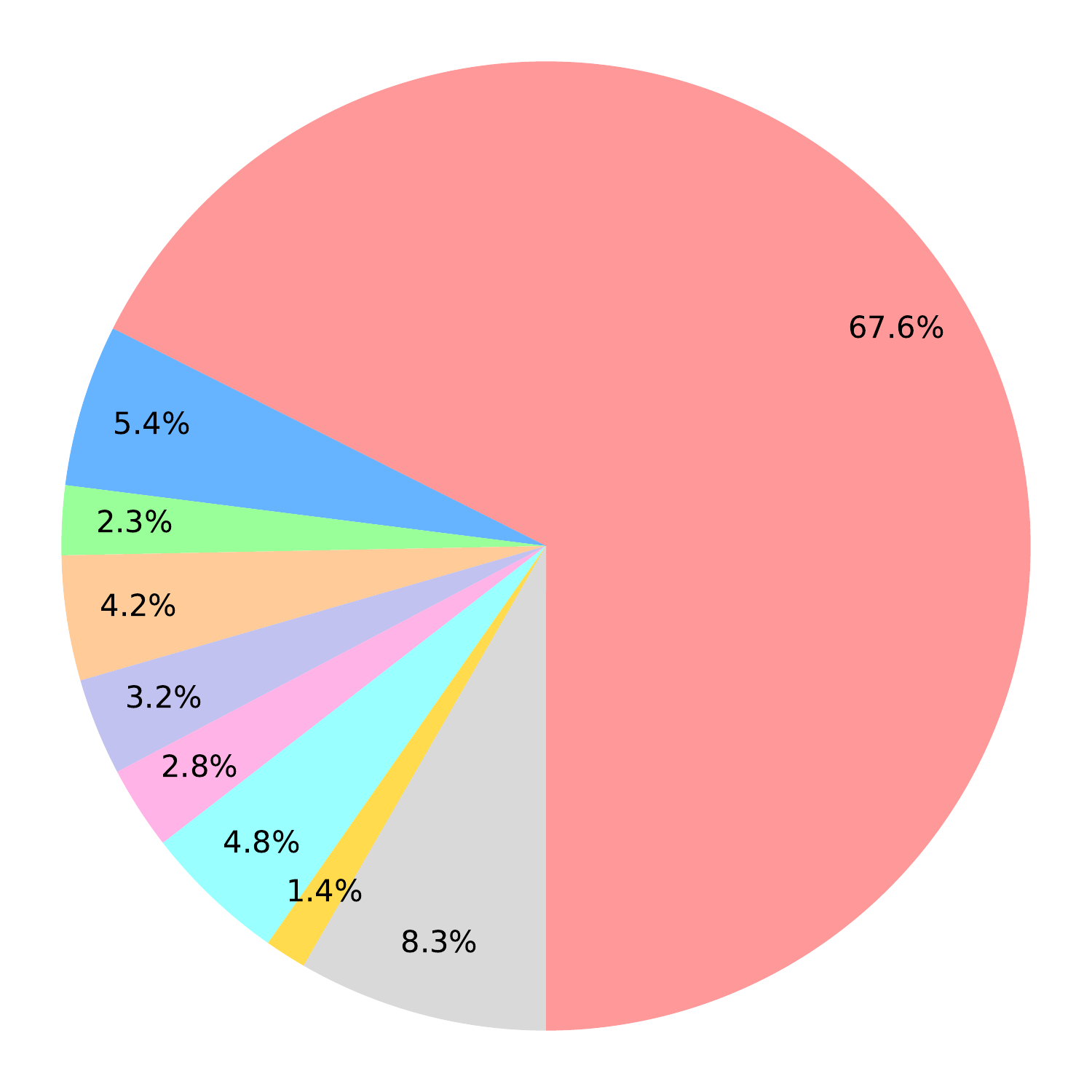}
        \caption*{$\pi_{\theta, \text{RL}}^1 \rightrightarrows $ initial dataset.}
    \end{minipage}
    \hfill
    \begin{minipage}[b]{0.32\textwidth}
        \includegraphics[width=\linewidth]{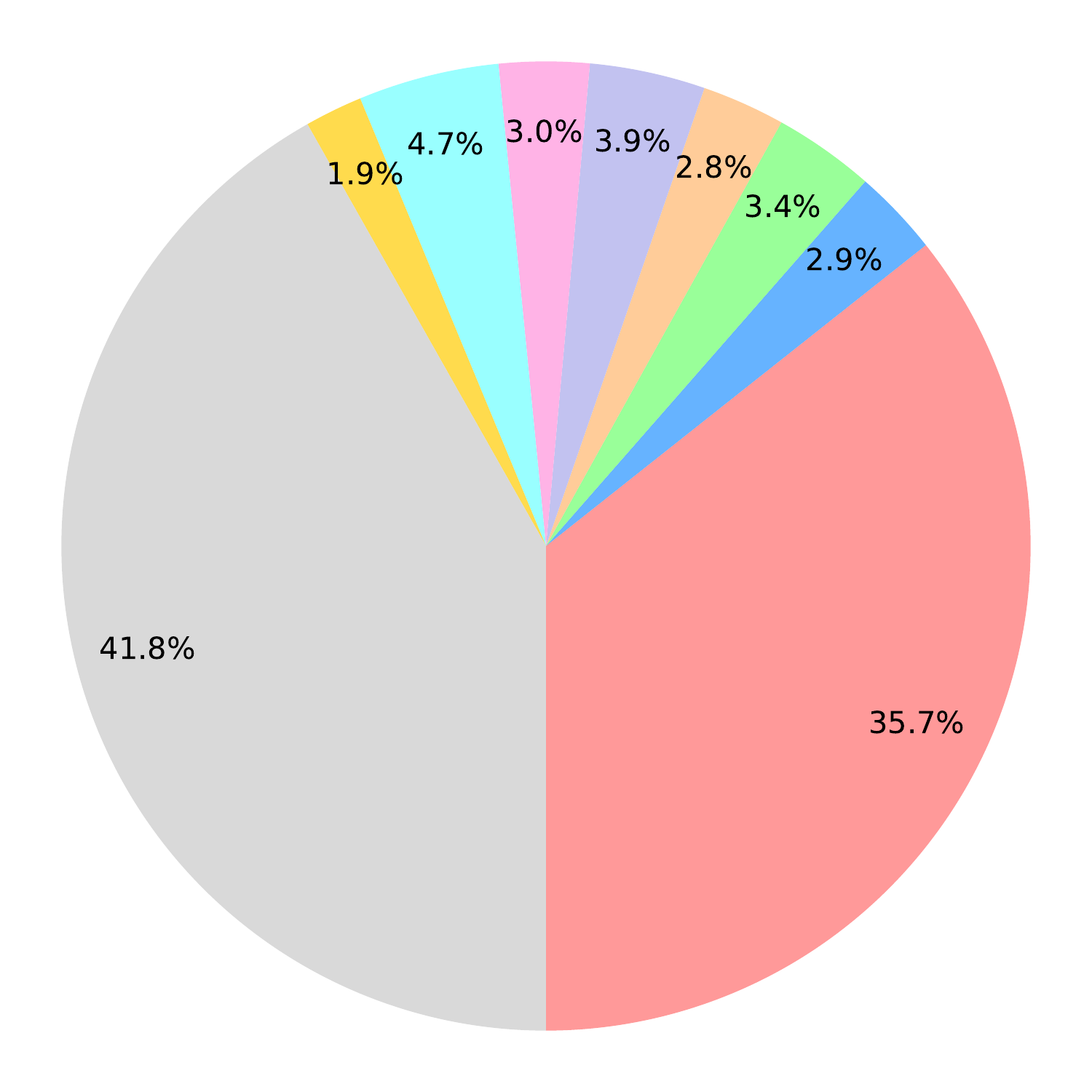}
        \caption*{$\pi_{\theta, \text{RL}}^1 \rightrightarrows  1^{st}$ dataset.}
    \end{minipage}
    \hfill
    \begin{minipage}[b]{0.32\textwidth}
        \centering
        \includegraphics[height=4cm, keepaspectratio]{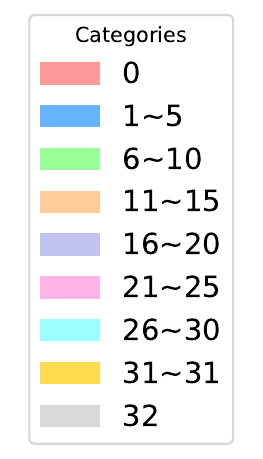}
        \caption*{Legend}
    \end{minipage}

    \bigskip
    \begin{minipage}[b]{0.32\textwidth}
        \includegraphics[width=\linewidth]{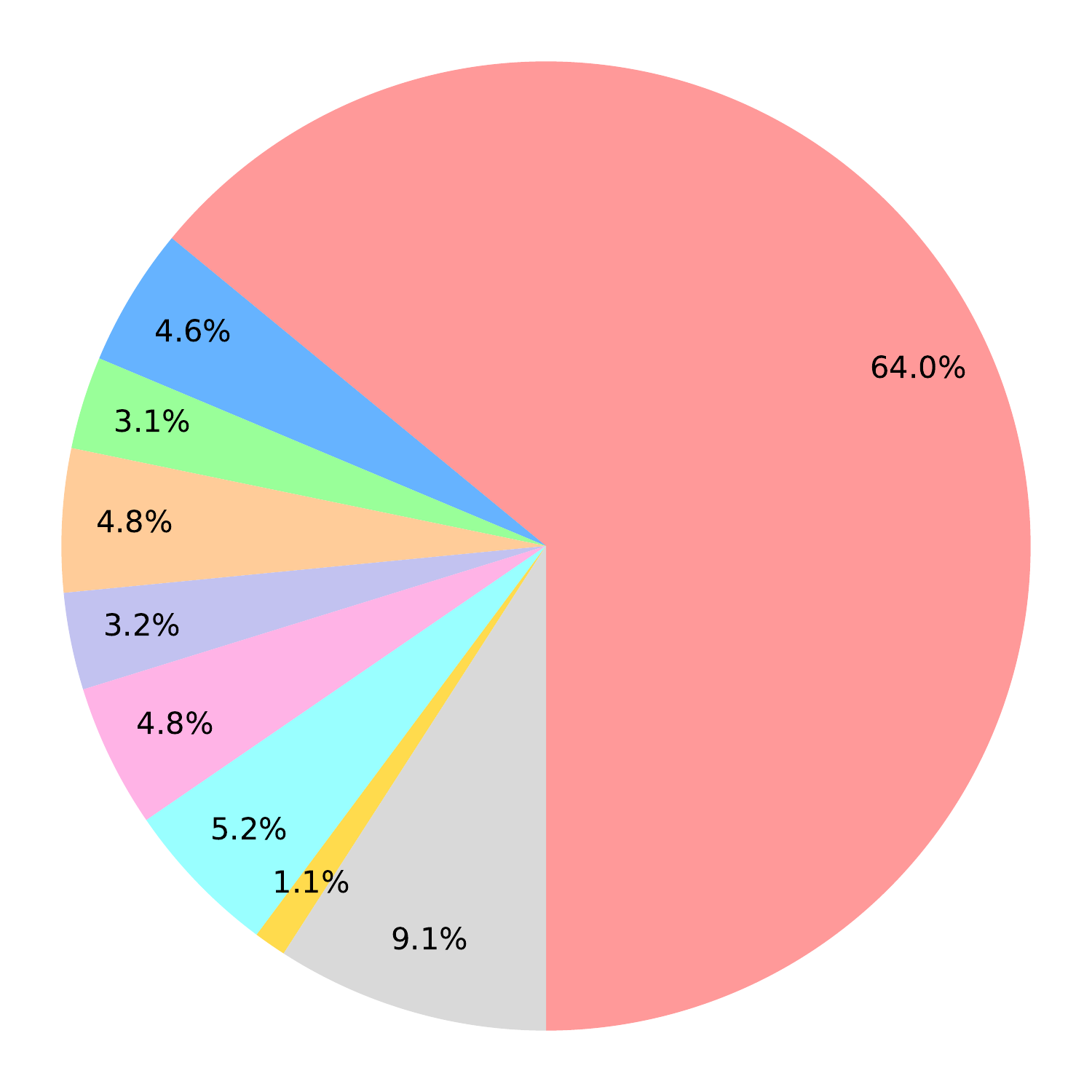}
        \caption*{$\pi_{\theta, \text{RL}}^2 \rightrightarrows $ initial dataset.}
    \end{minipage}
    \hfill
    \begin{minipage}[b]{0.32\textwidth}
        \includegraphics[width=\linewidth]{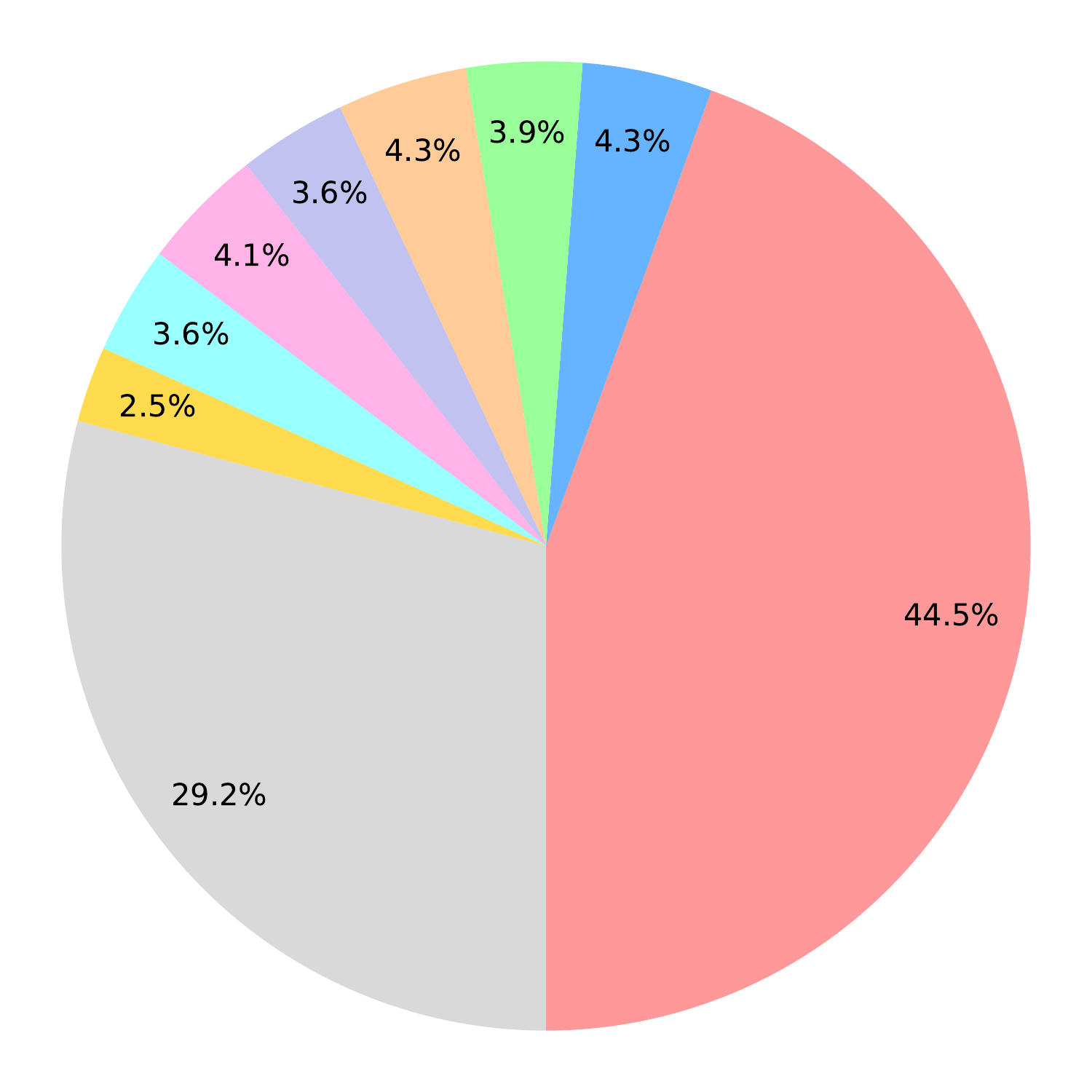}
        \caption*{$\pi_{\theta, \text{RL}}^2 \rightrightarrows  1^{st}$ dataset.}
    \end{minipage}
    \hfill
    \begin{minipage}[b]{0.32\textwidth}
        \includegraphics[width=\linewidth]{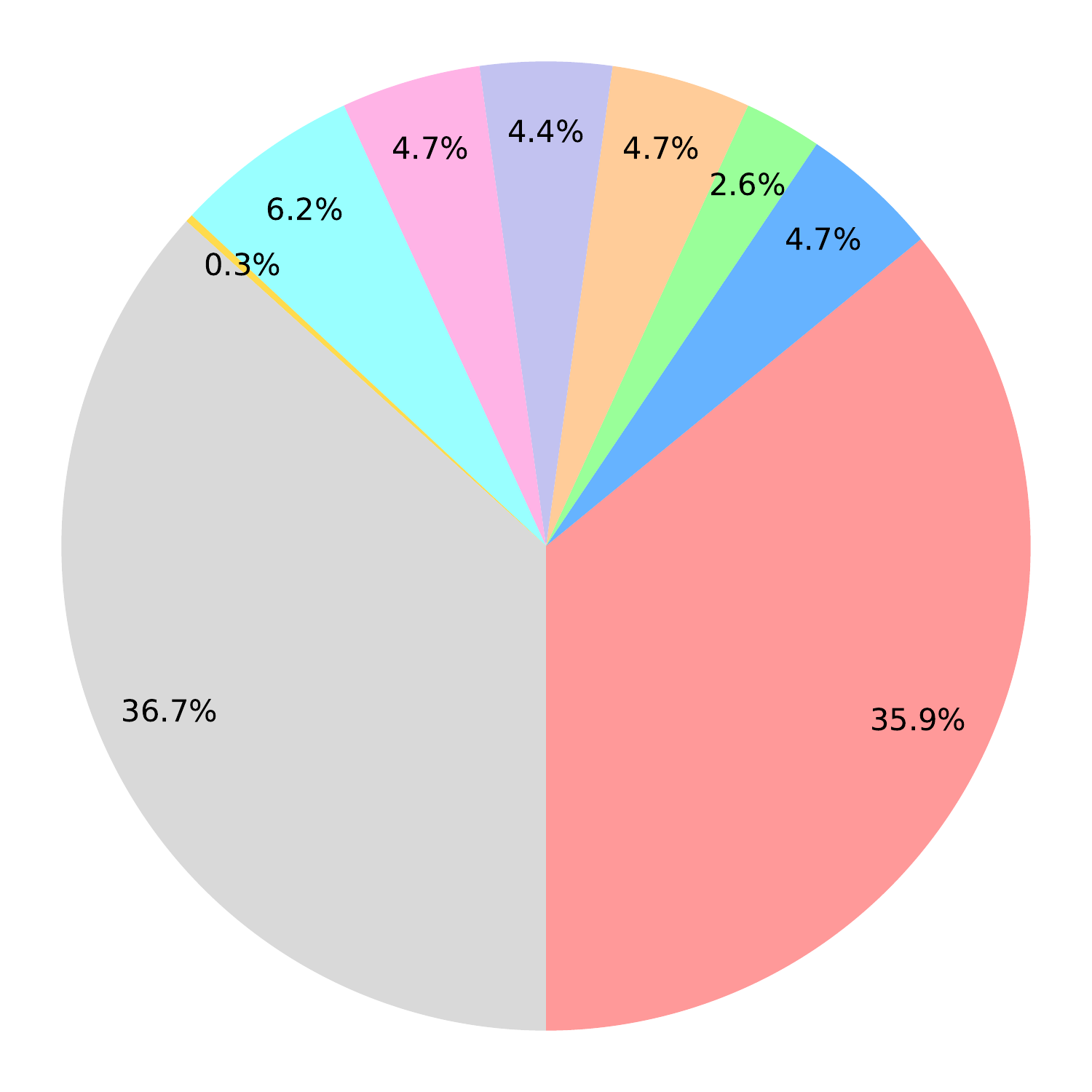}
        \caption*{$\pi_{\theta, \text{RL}}^2 \rightrightarrows  2^{nd}$ dataset.}
    \end{minipage}

    \caption{\textbf{Evolution of data difficulty across iterations (assessed by error-rate).} $ \pi_{\theta} \rightrightarrows \mathcal{D}$ denotes the difficulty distribtions of dataset $\mathcal{D}$ evaluated using model $\pi_{\theta}$ based on error-rate.}
    \label{fig:fig_error_rate_supple}
\end{figure*}

As shown in Figure \ref{fig:fig_error_rate_supple}, we also evaluate the difficulty of training data from previous iterations using the model training in the preceding iteration. Our observations are as follows:
1) In the evaluation of the second iteration, the first iteration model $\pi_{\theta, \text{RL}}^1$ achieved completely wrong predictions on approximately $8\%$ of the data from the first iteration.
2) Despite the fact that the data generated in the second iteration was more challenging than that of the first iteration (\cf, as evidenced by the longer reasoning length in the Figure $3$, the first iteration model $\pi_{\theta, \text{RL}}^1$ was still able to solve about $30\%$ of the problems from the second iteration.
3) In the evaluation of the third iteration, when the second iteration model $\pi_{\theta, \text{RL}}^2$ was used to evaluate the data from the first iteration, there is an increase in completely wrong predictions and a decrease in fully correct ones. This suggests that the model begins to exhibit \textit{catastrophic forgetting}.
4) Even within the second iteration data, around $30\%$ of the questions are completely wrong ,which aligns with the assumption stated in the main text that the data difficulty increases across iterations.


\subsection{The influence of different iteration strategies.} 

\begin{table}[]
\centering
\caption{Comparison of SFT vs. RL on first iteration.
}
\label{tab:tab_sft_rl_supple}
\scalebox{0.99}{
\begin{tabular}{lcccc}
        \toprule
        Iteration              & Model & SFT & RL & Geo-Sub-Aux \\ \midrule
        \multirow{3}{*}{First} &  $\pi_{\theta}^0$      &  \cmark   &    &   45.4          \\
                               & $\pi_{\theta}^0$       &     &  \cmark  &  44.7           \\ 
                             & $\pi_{\theta}^0$       &  \cmark   & \cmark   & 48.7            \\ 
                               \bottomrule
        \end{tabular}}
\end{table}

In this main paper, we present a comparison of different iteration strategies across the second and third iterations. To provide a more comprehensive evaluation, we further conduct experiments on different strategies in the first iteration, as shown in Table \ref{tab:tab_sft_rl_supple}. 
Under the same parameter settings, the results show that SFT+RL achieves the best performance in the first iteration, followed by SFT alone, while using RL-only training yields the least favorable outcomes.

\newpage
\section*{NeurIPS Paper Checklist}

\begin{enumerate}

\item {\bf Claims}
    \item[] Question: Do the main claims made in the abstract and introduction accurately reflect the paper's contributions and scope?
    \item[] Answer: \answerYes{} 
    \item[] Justification: The abstract and introduction in this paper accurately reflect the paper's made in the paper.

\item {\bf Limitations}
    \item[] Question: Does the paper discuss the limitations of the work performed by the authors?
    \item[] Answer: \answerYes{} 
    \item[] Justification: We discuss the limitations of the work in Conclusion \ref{sec:conclusion}

\item {\bf Theory assumptions and proofs}
    \item[] Question: For each theoretical result, does the paper provide the full set of assumptions and a complete (and correct) proof?
    \item[] Answer: \answerYes{} 
    \item[] Justification: In this paper, all proofs of theorems are provided and all assumptions are clearly stated or referenced in the statement of any theorems.

    \item {\bf Experimental result reproducibility}
    \item[] Question: Does the paper fully disclose all the information needed to reproduce the main experimental results of the paper to the extent that it affects the main claims and/or conclusions of the paper (regardless of whether the code and data are provided or not)?
    \item[] Answer: \answerYes{} 
    \item[] Justification: All the results in this paper can be reproduced.

\item {\bf Open access to data and code}
    \item[] Question: Does the paper provide open access to the data and code, with sufficient instructions to faithfully reproduce the main experimental results, as described in supplemental material?
    \item[] Answer: \answerYes{} 
    \item[] Justification: We will release the code after the paper is accepted.

\item {\bf Experimental setting/details}
    \item[] Question: Does the paper specify all the training and test details (e.g., data splits, hyperparameters, how they were chosen, type of optimizer, etc.) necessary to understand the results?
    \item[] Answer:  \answerYes{} 
    \item[] Justification: The paper specify all the training and test details necessary to understand the results.

\item {\bf Experiment statistical significance}
    \item[] Question: Does the paper report error bars suitably and correctly defined or other appropriate information about the statistical significance of the experiments?
    \item[] Answer: \answerYes{} 
    \item[] Justification: We report error bars suitably and correctly defined or other appropriate information about the statistical significance of the experiments in this paper

\item {\bf Experiments compute resources}
    \item[] Question: For each experiment, does the paper provide sufficient information on the computer resources (type of compute workers, memory, time of execution) needed to reproduce the experiments?
    \item[] Answer:  \answerYes{} 
    \item[] Justification: For each experiment, the paper provide sufficient information on the computer resources needed to reproduce the experiments.
    
\item {\bf Code of ethics}
    \item[] Question: Does the research conducted in the paper conform, in every respect, with the NeurIPS Code of Ethics \url{https://neurips.cc/public/EthicsGuidelines}?
    \item[] Answer:  \answerYes{} 
    \item[] Justification: The research conducted in the paper conform, in every respect, with the NeurIPS Code of Ethics.

\item {\bf Broader impacts}
    \item[] Question: Does the paper discuss both potential positive societal impacts and negative societal impacts of the work performed?
    \item[] Answer:\answerYes{} 
    \item[] Justification: The paper discuss both potential positive societal impacts and negative societal impacts of the work performed in Appendix.
    
\item {\bf Safeguards}
    \item[] Question: Does the paper describe safeguards that have been put in place for responsible release of data or models that have a high risk for misuse (e.g., pretrained language models, image generators, or scraped datasets)?
    \item[] Answer: \answerNA{}
    \item[] Justification: The paper poses no such risks.

\item {\bf Licenses for existing assets}
    \item[] Question: Are the creators or original owners of assets (e.g., code, data, models), used in the paper, properly credited and are the license and terms of use explicitly mentioned and properly respected?
    \item[] Answer: \answerYes{}
    \item[] Justification: The creators or original owners of assets used in the paper are properly credited and the license and terms of use are explicitly mentioned and properly respected.

\item {\bf New assets}
    \item[] Question: Are new assets introduced in the paper well documented and is the documentation provided alongside the assets?
    \item[] Answer: \answerNA{} 
    \item[] Justification: The paper does not release new assets.

\item {\bf Crowdsourcing and research with human subjects}
    \item[] Question: For crowdsourcing experiments and research with human subjects, does the paper include the full text of instructions given to participants and screenshots, if applicable, as well as details about compensation (if any)? 
    \item[] Answer: \answerNA{}
    \item[] Justification:  The paper does not involve crowdsourcing nor research with human subjects.

\item {\bf Institutional review board (IRB) approvals or equivalent for research with human subjects}
    \item[] Question: Does the paper describe potential risks incurred by study participants, whether such risks were disclosed to the subjects, and whether Institutional Review Board (IRB) approvals (or an equivalent approval/review based on the requirements of your country or institution) were obtained?
    \item[] Answer: \answerNA{} 
    \item[] Justification: The paper does not involve crowdsourcing nor research with human subjects.

\item {\bf Declaration of LLM usage}
    \item[] Question: Does the paper describe the usage of LLMs if it is an important, original, or non-standard component of the core methods in this research? Note that if the LLM is used only for writing, editing, or formatting purposes and does not impact the core methodology, scientific rigorousness, or originality of the research, declaration is not required.
    \item[] Answer: \answerYes{} 
    \item[] Justification: We have provided a detailed description of the usage of LLMs in the paper.

\end{enumerate}

\end{document}